\documentclass[runningheads]{llncs}

\usepackage{graphicx}
\usepackage{comment}
\usepackage{amsmath,amssymb}
\usepackage{color}
\usepackage{url}
\usepackage{hyperref}
\usepackage[flushleft]{threeparttable}
\usepackage{gensymb}

\usepackage{amsmath}
\usepackage{amsfonts}
\usepackage{graphicx}
\usepackage[tight,footnotesize]{subfigure}
\usepackage{multirow}
\usepackage{color}
\usepackage{url}
\usepackage[affil-it]{authblk}
\usepackage{wrapfig}

\usepackage[T1]{fontenc}
\usepackage{times}
\usepackage[table]{xcolor}
\usepackage{graphicx}
\usepackage{enumitem}
\usepackage{booktabs}
\usepackage{wrapfig}
\usepackage{makecell}
\usepackage{graphicx}

\newif\ifreview
\reviewtrue
\reviewfalse

\ifreview
	\usepackage{lineno}

	\linenumbers
\fi

\begin{document}


\def\SubNumber{083}

\def\GCPRTrack{Regular Track}

\title{Implicit and Explicit Attention for Zero-Shot Learning}

\newcommand{\anjan}[1]{\textcolor{red}{Anjan: #1}}
\newcommand{\myparagraph}[1]{\vspace{1pt}\noindent{\bf #1}}

\ifreview
\else

	\author{Faisal Alamri\orcidID{0000-0001-6695-2504} \and
	Anjan Dutta\orcidID{0000-0002-1667-2245}}
	\authorrunning{F. Alamri and A. Dutta}
	\institute{University of Exeter, Streatham Campus, Exeter, EX4 4RN, United Kingdom\\
	\email{\{F.Alamri2,A.Dutta\}@exeter.ac.uk}}
\fi

\maketitle              

\begin{abstract}
Most of the existing Zero-Shot Learning (ZSL) methods focus on learning a compatibility function between the image representation and class attributes. Few others concentrate on learning image representation combining local and global features. However, the existing approaches still fail to address the bias issue towards the seen classes. In this paper, we propose implicit and explicit attention mechanisms to address the existing bias problem in ZSL models. We formulate the implicit attention mechanism with a self-supervised image angle rotation task, which focuses on specific image features aiding to solve the task. The explicit attention mechanism is composed with the consideration of a multi-headed self-attention mechanism via Vision Transformer model, which learns to map image features to semantic space during the training stage. We conduct comprehensive experiments on three popular benchmarks: AWA2, CUB and SUN. The performance of our proposed attention mechanisms has proved its effectiveness, and has achieved the state-of-the-art harmonic mean on all the three datasets.


\end{abstract}

%
%
%

\section{Introduction}
\label{sec:intro}
Most of the existing Zero-Shot Learning (ZSL) methods \cite{CLSWGAN,Schnfeld2019GeneralizedZA} depend on pretrained visual features and necessarily focus on learning a compatibility function between the visual features and semantic attributes. Recently, attention-based approaches have got a lot of popularity, as they allow to obtain an image representation by directly recognising object parts in an image that correspond to a given set of attributes \cite{AREN,S2GA}. Therefore, models capturing global and local visual information have been quite successful \cite{AREN,APN}. Although visual attention models quite accurately focus on object parts, it has been observed that often recognised parts in image and attributes are biased towards training (or \emph{seen}) classes due to the learned correlations \cite{APN}. This is mainly because the model fails to decorrelate the visual attributes in images.

\begin{figure} [!t]
\centering
\includegraphics[width=\linewidth]{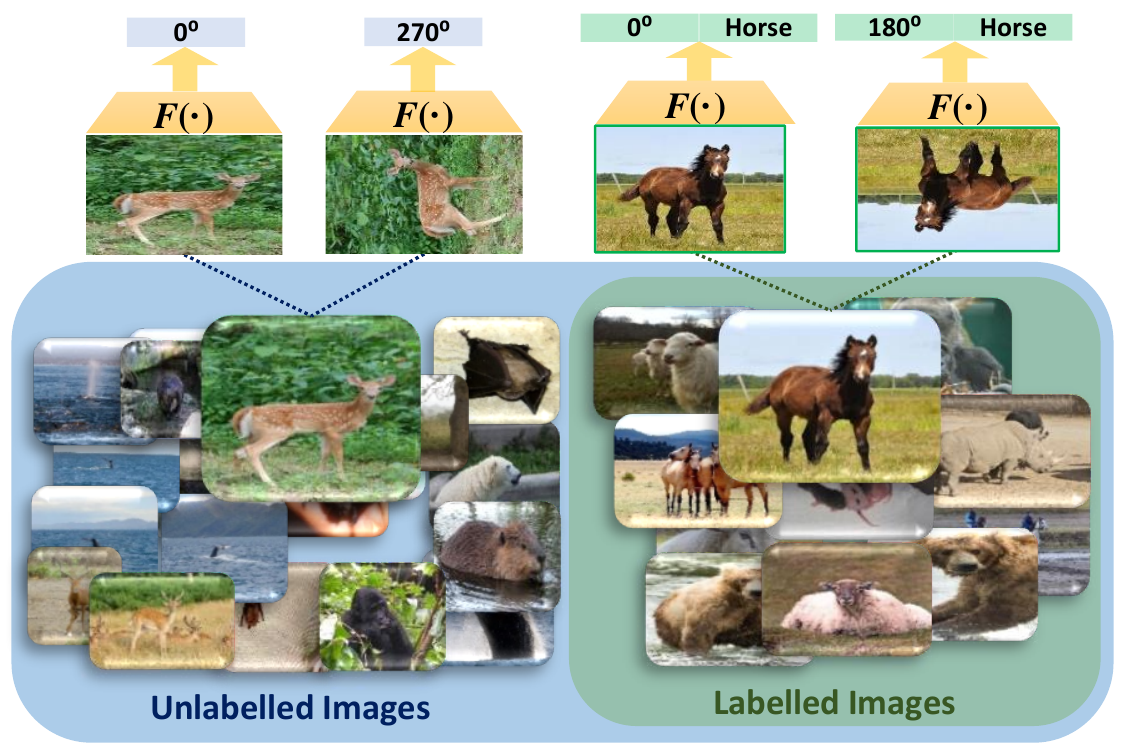}
\caption{Our method maps the visual features to the semantic space provided with two different input images (unlabelled and labelled data). Green represents the labelled images provided to train the model to capture visual features and predict object classes. Blue represents the unlabelled images that are rotated and attached to the former set of images to recognise rotated image angles in a self-supervised task. The model learns the visual representations of the rotated images implicitly via the use of attention. (Best viewed in colour)}
\label{fig:Model_into}
\end{figure}

Therefore, to alleviate these difficulties, in this paper, we consider two alternative attention mechanisms for reducing the effect of bias towards training classes in ZSL models. The first mechanism is via the self-supervised pretext task, which implicitly attends to specific parts of an image to solve the pretext task, such as recognition of image rotation angle \cite{Gidaris2018UnsupervisedRL}. For solving the pretext task, the model essentially focuses on learning image features that lead to solving the pretext task. Specifically, in this work, we consider rotating the input image concurrently by four different angles $(0^\circ,90^\circ,180^\circ,270^\circ)$ and then predicting the rotation class. Since pretext tasks do not involve attributes or class-specific information, the model does not learn the correlation between visual features and attributes. Our second mechanism employs the Vision Transformer (ViT) \cite{ViT} for mapping the visual features to semantic space. ViT having a rich multi-headed self-attention mechanism explicitly attends to those image parts related to class attributes. In a different setting, we combine the implicit with the explicit attention mechanism to learn and attend to the necessary object parts in a decorrelated or independent way. We attest that incorporating the rotation angle recognition in a self-supervised approach with the use of ViT does not only improve the ZSL performance significantly, but also and more importantly, contributes to reducing the bias towards seen classes, which is still an open challenge in the Generalised Zero-Shot Learning (GZSL) task \cite{32933444718}. Explicit use of attention mechanism is also examined, where the model is shown to enhance the visual feature localisation and attends to both global and discriminative local features guided by the semantic information given during training. As illustrated in Fig. \ref{fig:Model_into}, images fed into the model are taken from two different sources: 1) labelled images, which are the training images taken from the \emph{seen} classes, shown in green colour, and 2) other images, which could be taken from any source, shown in blue. The model is donated as (\(\mathcal{F(.)}\)), in this paper, we implement \(\mathcal{F(.)}\) either by ViT or by ResNet-101 \cite{He2016DeepRL} backbones. The first set of images is used to train the model to predict class attributes leading to the class labels via nearest search. However, the second set of images is used for rotation angle recognition, guiding the model to learn visual representations via implicit attention mechanism.

To summarise, in this paper, we make the following contributions: (1) We propose the utilisation of alternative attention mechanisms for reducing the bias towards the seen classes in zero-shot learning. By involving self-supervised pretext task, our model implicitly attends decorrelated image parts aiding to solve the pretext task, which learns image features independent of the training classes. (2) We perform extensive experiments on three challenging benchmark datasets, i.e. AWA2, CUB and SUN, in the generalised zero-shot learning setting and demonstrate the effectiveness of our proposed alternative attention mechanisms. We also achieve consistent improvement over the state-of-the-art methods. (3) The proposed method is evaluated with two backbone models: ResNet-101 and ViT, and shows significant improvement in the model performances, and reduces the issue of bias towards seen classes. We also show the effectiveness of our model qualitatively by plotting the attention maps. 




\section{Related Work} \label{sec:related}
In this section we briefly review the related arts on zero-shot learning, Vision Transformer and self-supervised learning.

\myparagraph{Zero-Shot Learning (ZSL):} Zero-Shot Learning (ZSL) uses semantic side information such as attributes and word embeddings \cite{AWA2:journals/corr/XianSA17,Mikolov2013DistributedRO,Pennington2014GloveGV,Dutta2020SEMPCYCIJCV,Federici2020MBI,Alamri2021ViTZSL} to predict classes that have never been presented during training. Early ZSL models train different attribute classifiers assuming independence of attributes and then estimate the posterior of the test classes by combining attribute prediction probabilities \cite{5206594}. Others do not follow the independence assumption and learn a linear \cite{DeViSE,SJE,ALE} or non-linear \cite{LATEM} compatibility function from visual features to semantic space. There are some other works that learn an inverse mapping from semantic to visual feature space \cite{Shigeto2015RidgeRH,Zhang2017LearningAD}. Learning a joint mapping function for each space into a common space (i.e. a shared latent embedding) is also investigated in \cite{LATEM,8237715,Gune2018StructureAD}. Different from the above approaches, generative models synthesise samples of unseen classes based on information learned from seen classes and their semantic information, to tackle the issue of bias towards the seen classes \cite{CLSWGAN,GAZSL,Schnfeld2019GeneralizedZA}. Unlike other models, which focus on the global visual features, attention-based methods aim to learn discriminative local visual features and then combine with the global information \cite{S2GA,SGMA}. Examples include S$^2$GA \cite{S2GA} and AREN \cite{AREN} that apply an attention-based network to incorporate discriminative regions to provide rich visual expression automatically. In addition, GEN \cite{Xie2020RegionGE} proposes a graph reasoning method to learn relationships among multiple image regions. Others focus on improving localisation by adapting the human gaze behaviour \cite{Liu2021GoalOrientedGE}, exploiting a global average pooling scheme as an aggregation mechanism \cite{SELAR} or by jointly learning both global and local features \cite{SGMA}. Inspired by the success of the recent attention-based ZSL models, in this paper, we propose two alternative attention mechanisms to capture robust image features suitable to ZSL task. Our first attention mechanism is implicit and is based on self-supervised pretext task \cite{Gidaris2018UnsupervisedRL}, whereas the second attention mechanism is explicit and is based on ViT \cite{ViT}. To the best of our knowledge, both of these attention models are still unexplored in the context of ZSL. Here we also point out that the inferential comprehension of visual representations upon the use of SSL and ViT is a future direction to consider for ZSL task. 








\myparagraph{Vision Transformer (ViT):} The Transformer \cite{Transfomer} adopts the self-attention mechanism to weigh the relevance of each element in the input data. Inspired by its success, it has been implemented to solve many computer vision tasks \cite{Alamri,ViT,khan2021transformers} and many enhancements and modifications of Vision Transformer (ViT) have been introduced. For example, CaiT \cite{Deep_transfomer} introduces deeper transformer networks, Swin Transformer \cite{Swin} proposes a hierarchical Transformer capturing visual representation by computing self-attention via shifted windows, and TNT \cite{TnT} applies the Transformer to compute the visual representations using both patch-level and pixel-level information. In addition, CrossViT \cite{CrossViT} proposes a dual-branch Transformer with different sized image patches. Recently, TransGAN \cite{TransGAN} proposes a completely free of convolutions generative adversarial network solely based on pure transformer-based architectures. Readers are referred to \cite{khan2021transformers}, for further reading about ViT based approaches. The applicability of ViT-based models is growing, but it has remained relatively unexplored to the zero-shot image recognition tasks where attention based models have already attracted a lot of attention. Therefore employing robust attention based models, such as ViT is absolutely timely and justified for improving the ZSL performance.


\myparagraph{Self-Supervised Learning (SSL):} Self-Supervised Learning (SSL) is widely used for unsupervised representation learning to obtain robust representations of samples from raw data without expensive labels or annotations. Although the recent SSL methods use contrastive objectives \cite{Chen2020SimCLR,Grill2020BYOL}, early works used to focus on defining pretext tasks, which typically involves defining a surrogate task on a domain with ample weak supervision labels, such as predicting the rotation of images \cite{Gidaris2018UnsupervisedRL}, relative positions of patches in an image \cite{DoerschGE15,NorooziF16}, image colours \cite{LarssonMS16,ZhangIE16} etc. Encoders trained to solve such pretext tasks are expected to learn general features that might be useful for other downstream tasks requiring expensive annotations (e.g. image classification). Furthermore, SSL has been widely used in various applications, such as few-shot learning \cite{Gidaris_2019_ICCV}, domain generalisation \cite{Carlucci2019JigsawDomGen} etc. In contrast, in this paper, we utilise the self-supervised pretext task of image rotation prediction for obtaining implicit image attention to solve ZSL.


\begin{figure} [!t]
\centering
\includegraphics[width=\linewidth, height=.58\textwidth]{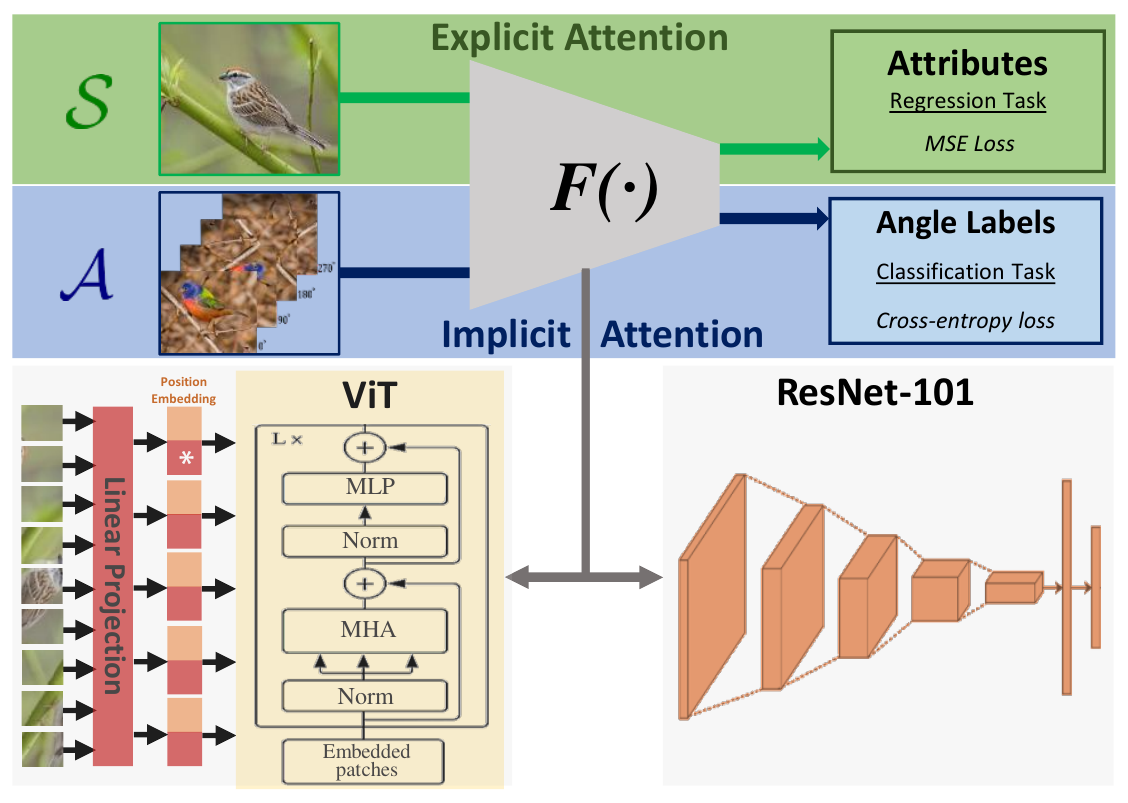}
\caption{IEAM-ZSL Architecture. IEAM-ZSL consists of two pipelines represented in Green and Blue colours, respectively. The former takes images from the ZSL datasets with their class-level information input to the Transformer Encoder for attributes predictions. Outputs are compared with semantic information of the corresponding images using MSE loss as a regression task. The latter, shown in Blue colour, is fed with images after generating four rotations for each (i.e. $0^{\circ}$, $90^{\circ}$, $180^{\circ}$, and $270^{\circ}$), to predict the rotation angle. At inference, solely the ZSL test datasets, with no data augmentation, are inputted to the model to predict the class-level attributes. A search for the nearest class label is then conducted.}
\label{fig:Model}
\end{figure}

\section{Implicit and Explicit Attention for Zero-Shot Learning}
\label{ProposedModel}

In this work, we propose an Implicit and Explicit Attention mechanism-based Model for solving image recognition in Zero-Shot Learning (IEAM-ZSL). We utilise self-supervised pretext tasks, such as image rotation angle recognition, for obtaining image attention in an implicit way. Here the main rational is for predicting the correct image rotation angle, the model needs to focus on image features with discriminative textures, colours etc., which implicitly attend to specific regions in an image. For having explicit image attention, we utilise the multi-headed self-attention mechanism involved in Vision Transform model.

From ZSL perspective, we follow the inductive approach for training our model, i.e. during training, the model only has access to the training set (\emph{seen} classes), consisting of only the labelled images and continuous attributes of the seen classes \((\mathcal{S}= \{\mathbf{x}, \mathbf{y} | \mathbf{x}\in \mathcal{X}, \mathbf{y} \in \mathcal{Y}^s \})\). An RGB image in image space $\mathcal{X}$ is denoted as $\mathbf{x}$, where $\mathbf{y}\in\mathcal{Y}$ is the class-level semantic vector annotated with $M$ different attributes. As depicted in Fig. \ref{fig:Model}, a $224 \times 224$ image $\mathbf{x} \in \mathbb{R}^{H \times W \times C}$ with resolution $H \times W$ and $C$ channels is fed into the model. Addition to $\mathcal{S}$, we also use an auxiliary set of unlabelled images $\mathcal{A} = \{\mathbf{x} \in \mathcal{X}\}$ for predicting the image rotation angle to obtain implicit attention. Note, here the images from $\mathcal{S}$ and $\mathcal{A}$ may or may not overlap, however, the method does not utilise the categorical or semantic label information of the images from the set $\mathcal{A}$.


\subsection{Implicit Attention}
Self-supervised pretext tasks provide a surrogate supervision signal for feature learning without any manual annotations \cite{Gidaris2018UnsupervisedRL,Dosovitskiy2014DiscriminativeUF,7410370} and it is well known that this type of supervision focuses on image features that help to solve the considered pretext task. It has also been shown that these pretext tasks focus on meaningful image features and effectively avoid learning correlation between visual features \cite{Gidaris2018UnsupervisedRL}. As self-supervised learning avoids considering semantic class labels, spurious correlation among visual features are not learnt. Therefore, motivated by the above facts, we employ an image rotation angle prediction task to obtain implicitly attended image features. For that, we rotate an image by $0^\circ$, $90^\circ$, $180^\circ$ and $270^\circ$, and train the model to correctly classify the rotated images. Let $g(\cdot|a)$ be an operator that rotates an image $\mathbf{x}$ by an angle $90^\circ \times a$, where $a\in\{0,1,2,3\}$. Now let $\hat{\mathbf{y}}_a$ be the predicted probability for the rotated image $\mathbf{x}_a$ with label $a$, then the loss for training the underlying model is computed as follows:
\begin{equation}
\mathcal{L}_\text{CE} = \displaystyle -\sum_{a=1}^{4} \log(\mathbf{\hat{y}}_a) 
\label{Eq11}
\end{equation}
In our case, the task of predicting image rotation angle trains the model to focus on specific image regions having rich visual features (for example, textures or colours). This procedure implicitly learns to attend image features.

\subsection{Explicit Attention}

For obtaining explicit attention, we employ Vision Transformer model \cite{ViT}, where each image $\mathbf{x} \in \mathbb{R}^{H \times W \times C}$ with resolution $H \times W$ and $C$ channels is fed into the model after resizing it to $224 \times 224$. Afterwards, the image is split into a sequence of $N$ patches denoted as $\mathbf{x}_p \in \mathbb{R}^{N \times (P^2.C)}$, where \(N = \frac{H.W}{P^2}\). Patch embeddings (small red boxes in Fig. \ref{fig:Model}) are encoded by applying a trainable 2D convolution layer with kernel size=(16, 16) and stride=(16, 16)). An extra learnable classification token ($\mathbf{z}^0_0 = \mathbf{x}_\text{class}$) is appended at the beginning of the sequence to encode the global image representation, which is donated as ($*$). Position embeddings (orange boxes) are then attached to the patch embeddings to obtain the relative positional information. Patch embeddings ($\mathbf{z}$) are then projected through a linear projection $\mathbf{E}$ to $D$ dimension (i.e. $D=1024$) as in Eq. \ref{Eq2.1}. Embeddings are then passed to the Transformer encoder, which consists of Multi-Head Attention (MHA) (Eq. \ref{Eq2.2}) and MLP blocks (Eq. \ref{Eq2.3}). A layer normalisation (Norm) is applied before every block, and residual connections after every block. The image representation ($\mathbf{\hat{y}}$) is then produced as in Eq. \ref{Eq2.4}.

\begin{align}
\mathbf{z}_0 &=[\mathbf{x}_\text{class}; \mathbf{x}^{1}_{p}\mathbf{E};\mathbf{x}^{2}_{p}\mathbf{E}; \ldots; \mathbf{x}^{N}_{p}\mathbf{E}] + \mathbf{E}_\text{pos}, & &\mathbf{E} \in \mathbb{R}^{(P^2.C) \times D}, \mathbf{E}_\text{pos} \in \mathbb{R}^{(N+1) \times D} \label{Eq2.1}\\
\mathbf{z}^{\prime}_{\ell} &= \text{MHA}(\text{Norm}(\mathbf{z}_{\ell-1})) + \mathbf{z}_{\ell-1}, & &\ell = 1 \ldots L\;(L=24) \label{Eq2.2}\\
\mathbf{z}_{\ell}&=\text{MLP}(\text{Norm}(\mathbf{z}^{\prime}_{\ell})) + \mathbf{z}^{\prime}_{\ell}, & &\ell = 1 \ldots L \label{Eq2.3}\\
\mathbf{\hat{y}} &=\text{Norm}(\mathbf{z}^0_L) \label{Eq2.4}
\end{align}
Below we provide details of our multi-head attention mechanism within the ViT model.

\myparagraph{Multi-Head Attention (MHA):} Patch embeddings are fed into the transformer encoder, where the multi-head attention takes place. Self-attention is performed for every patch in the sequence of the patch embeddings independently; thus, attention works simultaneously for all the patches, leading to multi-headed self-attention. This is computed by creating three vectors, namely Query ($Q$), Key ($K$) and Value ($V$). They are created by multiplying the patch embeddings by three trainable weight matrices (i.e. $W^Q$, $W^K$ and $W^V$) applied to compute the self-attention. A dot-product operation is performed on the $Q$ and $K$ vectors, calculating a scoring matrix that measures how much a patch embedding has to attend to every other patch in the input sequence. The score matrix is then scaled down and converted into probabilities using a softmax. Probabilities are then multiplied by the $V$ vectors, as in Eq. \ref{Eq2.5}, where \textit{$d_k$} is the dimension of the vector $K$. Multi-headed self-attention mechanism produces a number of self-attention matrices which are concatenated and fed into a linear layer and passed sequentially to 1) regression head and 2) classification head.
\begin{equation}
\text{Attention(\textit{Q}, \textit{K}, \textit{V})}= \text{softmax}(\frac{QK^{T}}{\sqrt{d_k}})\text{\textit{V}}
\label{Eq2.5}
\end{equation}
The multi-headed self-attention mechanism involved in the Vision Transformer guides our model to learn both the global and local visual features. It is worth noting that the standard ViT has only one classification head implemented by an MLP, which is changed in our model to two heads to meet the two different underlying objectives. The first head is a regression head applied to predict $M$ different class attributes, whereas the second head is added for rotation angle classification. For the former task, the objective function employed is the Mean Squared Error (MSE) loss as in Eq. \ref{Eq1}, where $\mathbf{y}_i$ is the target attributes, and $\mathbf{\hat{y}}_i$ is the predicted ones. For the latter task, cross-entropy (Eq. \ref{Eq11}) objective is applied.
\begin{equation}
\mathcal{L}_\text{MSE} = \displaystyle\frac{1}{M}\sum_{i=1}^{M}(\mathbf{y}_i - \mathbf{\hat{y}}_i)^2 
\label{Eq1}
\end{equation}
The total loss used for training our model is defined in Eq. \ref{Eqlos}, where $\lambda_1 = 1$ and $\lambda_2 = 1$. 
\begin{equation}
\mathcal{L}_\text{TOT} = \lambda_1 \mathcal{L}_\text{CE} + \lambda_2 \mathcal{L}_\text{MSE}
\label{Eqlos}
\end{equation}
During the inference phase, original test images from the seen and unseen classes are inputted. Class labels are then determined using the cosine similarity between the predicted attributes and every target class embeddings predicted by our model. 


\section{Experiments} \label{Experiment}


\myparagraph{Datasets:} We have conducted our experiments on three popular ZSL datasets: AWA2, CUB, and SUN, whose details are presented in Table \ref{tab:mytab}. The main aim of this experimentation is to validate our proposed method IEAM-ZSL, demonstrating its effectiveness and comparing it with the existing state-of-the-art methods. Among these datasets, AWA2 \cite{AWA2:journals/corr/XianSA17} consists of $37,322$ images of $50$ categories ($40$ seen + $10$ unseen). Each category contains $85$ binary as well as continuous class attributes. CUB \cite{CUB} contains $11,788$ images forming $200$ different types of birds, among them $150$ classes are considered as seen, and the other $50$ as unseen, which is split by \cite{ALE}. Together with images CUB dataset also contains $312$ attributes describing birds. Finally, SUN \cite{SUN} has the largest number of classes among others. It consists of $717$ types of scene images, which divided into $645$ seen and $72$ unseen classes. The SUN dataset contains $14,340$ images with $102$ annotated attributes.


\begin{table}[!ht]
\begin{center}
\caption{Dataset statistics: The number of classes (seen + unseen classes shown within parenthesis), the number of attributes and the number of images per dataset.} \label{tab:mytab}
\begin{tabular}{|c|c|c|c|}
\hline
Datasets  & \cellcolor{red!10}AWA2 \cite{AWA2:journals/corr/XianSA17}  & \cellcolor{green!10}CUB \cite{CUB}  & \cellcolor{blue!10}SUN \cite{SUN} \\\hline
Number of Classes  & \cellcolor{red!10}$50$   & \cellcolor{green!10}$200$  &  \cellcolor{blue!10}$717$ \\
(Seen + Unseen) & \cellcolor{red!10} $(40 + 10)$  & \cellcolor{green!10}  $(150 + 50)$ &  \cellcolor{blue!10} $(645 + 72)$ \\\hline 
Number of Attributes & \cellcolor{red!10}$85$ & \cellcolor{green!10}$312$ & \cellcolor{blue!10}$102$ \\\hline
Number of Images & \cellcolor{red!10}$37,322$  & \cellcolor{green!10}$11,788$  & \cellcolor{blue!10}$14,340$ \\
\hline
\end{tabular}
\end{center}
\end{table}

\myparagraph{Implementation Details:} In our experiment, we have used two different backbones: (1) ResNet-101 and (2) Vision Transformer (ViT), both of which are pretrained on ImageNet and then finetuned for the ZSL tasks on the datasets mentioned above. We resize the image to $224 \times 224$ before inputting it into the model. For ViT, the primary baseline model employed uses an input patch size $16 \times 16$, with $1024$ hidden dimension, and having $24$ layers and $16$ heads on each layer, and $24$ series encoder. We use the Adam optimiser for training our model with a fixed learning rate of $0.0001$ and a batch size of $64$. In the setting where we use self-supervised pretext task, we construct the batch with $32$ \emph{seen} training images from set $\mathcal{S}$ and $32$ rotated images (i.e. eight images, where each image is rotated to $0^\circ$, $90^\circ$, $180^\circ$ and $270^\circ$) from set $\mathcal{A}$. We have implemented our model with PyTorch\footnote{Our code is available at: \url{https://github.com/FaisalAlamri0/IEAM-ZSL}} deep learning framework and trained the model on a GeForce RTX 3090 GPU on a workstation with Xeon processor and 32GB of memory.



\myparagraph{Evaluation:} The proposed model is evaluated on the three above mentioned datasets. We have followed the inductive approach for training our model, i.e. our model has no access to neither visual nor side-information of unseen classes during training. During the evaluation, we have followed the GZSL protocol. Following \cite{8413121}, we compute the top-1 accuracy for both seen and unseen classes. In addition, the harmonic mean of the top-1 accuracies on the seen and unseen classes is used as the main evaluation criterion. Inspired by the recent works \cite{SELAR,AREN,calibrated}, we have used the Calibrated Stacking \cite{calibrated} for evaluating our model under GZSL setting. The calibration factor $\gamma$ is dataset-dependent and decided based on a validation set. For AWA2 and CUB, the calibration factor $\gamma$ is set to $0.9$ and for SUN, it is set to $0.4$.





\subsection{Quantitative Results}
\label{sec:quant_results}

\begin{table}[!ht]
\centering
\caption{Generalised zero-shot classification performance on AWA2, CUB and SUN. Reported models are ordered in terms of their publishing dates. Results are reported in \%.}
\label{tab:Perfromance}
\begin{tabular}{|l|c|c|c|c|c|c|c|c|c|c|} 
\hline
\multirow{2}{*}{Models} 
& \multicolumn{3}{c|}{\cellcolor{red!10}AWA2} 
& \multicolumn{3}{c|}{\cellcolor{green!10}CUB} 
& \multicolumn{3}{c|}{\cellcolor{blue!10}SUN}   \\ 
\cline{2-10}
& \cellcolor{red!10}S & \cellcolor{red!10}U & \cellcolor{red!10}H 
& \cellcolor{green!10}S & \cellcolor{green!10}U & \cellcolor{green!10}H 
& \cellcolor{blue!10}S & \cellcolor{blue!10}U & \cellcolor{blue!10}H \\
\hline
DAP \cite{5206594}   & \cellcolor{red!10}84.7 & \cellcolor{red!10}0.0 & \cellcolor{red!10}0.0  & \cellcolor{green!10}67.9 & \cellcolor{green!10}1.7 & \cellcolor{green!10}3.3 & \cellcolor{blue!10}25.1 & \cellcolor{blue!10}4.2 & \cellcolor{blue!10}7.2 \\

IAP \cite{5206594}   & \cellcolor{red!10}87.6 & \cellcolor{red!10}0.9 & \cellcolor{red!10}1.8 & \cellcolor{green!10}72.8 & \cellcolor{green!10}0.2 & \cellcolor{green!10}0.4 & \cellcolor{blue!10}37.8 & \cellcolor{blue!10}1.0 & \cellcolor{blue!10}1.8 \\

DeViSE \cite{DeViSE} & \cellcolor{red!10}74.7 & \cellcolor{red!10}17.1 & \cellcolor{red!10}27.8 & \cellcolor{green!10}53.0 & \cellcolor{green!10}23.8 & \cellcolor{green!10}32.8 & \cellcolor{blue!10}30.5 & \cellcolor{blue!10}14.7 & \cellcolor{blue!10}19.8 \\

ConSE \cite{ConSE}   & \cellcolor{red!10}{\color{red}90.6} & \cellcolor{red!10}0.5 & \cellcolor{red!10}1.0 & \cellcolor{green!10}72.2 & \cellcolor{green!10}1.6 & \cellcolor{green!10}3.1 & \cellcolor{blue!10}39.9 & \cellcolor{blue!10}6.8 & \cellcolor{blue!10}11.6 \\

ESZSL \cite{romera-paredes15}   & \cellcolor{red!10}77.8 & \cellcolor{red!10}5.9 & \cellcolor{red!10}11.0 & \cellcolor{green!10}63.8 & \cellcolor{green!10}12.6 & \cellcolor{green!10}21.0 & \cellcolor{blue!10}27.9 & \cellcolor{blue!10}11.0 & \cellcolor{blue!10}15.8 \\

SJE \cite{SJE}       & \cellcolor{red!10}73.9 & \cellcolor{red!10}8.0 & \cellcolor{red!10}14.4 & \cellcolor{green!10}59.2 & \cellcolor{green!10}23.5 & \cellcolor{green!10}33.6 & \cellcolor{blue!10}30.5 & \cellcolor{blue!10}14.7 & \cellcolor{blue!10}19.8 \\

SSE \cite{SSE}       & \cellcolor{red!10}82.5 & \cellcolor{red!10}8.1 & \cellcolor{red!10}14.8 & \cellcolor{green!10}46.9 & \cellcolor{green!10}8.5 & \cellcolor{green!10}14.4 & \cellcolor{blue!10}36.4 & \cellcolor{blue!10}2.1 & \cellcolor{blue!10}4.0 \\

LATEM \cite{LATEM}   & \cellcolor{red!10}77.3 & \cellcolor{red!10}11.5 & \cellcolor{red!10}20.0 & \cellcolor{green!10}57.3 & \cellcolor{green!10}15.2 & \cellcolor{green!10}24.0 & \cellcolor{blue!10}28.8 & \cellcolor{blue!10}14.7 & \cellcolor{blue!10}19.5 \\

ALE \cite{ALE}       & \cellcolor{red!10}81.8 & \cellcolor{red!10}14.0 & \cellcolor{red!10}23.9 & \cellcolor{green!10}62.8 & \cellcolor{green!10}23.7 & \cellcolor{green!10}34.4 & \cellcolor{blue!10}33.1 & \cellcolor{blue!10}21.8 & \cellcolor{blue!10}26.3 \\

*GAZSL \cite{GAZSL}   & \cellcolor{red!10}86.5 & \cellcolor{red!10}19.2 & \cellcolor{red!10}31.4  & \cellcolor{green!10}60.6 & \cellcolor{green!10}23.9 & \cellcolor{green!10}34.3 & \cellcolor{blue!10}34.5 & \cellcolor{blue!10}21.7 & \cellcolor{blue!10}26.7 \\

SAE \cite{SAE}       & \cellcolor{red!10}82.2 & \cellcolor{red!10}1.1 & \cellcolor{red!10}2.2 & \cellcolor{green!10}54.0 & \cellcolor{green!10}7.8 & \cellcolor{green!10}13.6 & \cellcolor{blue!10}18.0 & \cellcolor{blue!10}8.8 & \cellcolor{blue!10}11.8 \\

*f-CLSWGAN \cite{CLSWGAN} & \cellcolor{red!10}64.4 & \cellcolor{red!10}{\color{red}57.9} & \cellcolor{red!10}59.6  & \cellcolor{green!10}57.7 & \cellcolor{green!10}43.7 & \cellcolor{green!10}49.7 & \cellcolor{blue!10}36.6 & \cellcolor{blue!10}42.6 & \cellcolor{blue!10}39.4 \\

AREN \cite{AREN}     & \cellcolor{red!10}79.1 & \cellcolor{red!10}54.7 & \cellcolor{red!10}64.7  & \cellcolor{green!10}{63.2} & \cellcolor{green!10}\color{blue}69.0 & \cellcolor{green!10}66.0 & \cellcolor{blue!10}40.3 & \cellcolor{blue!10}32.3 & \cellcolor{blue!10}35.9 \\

*f-VAEGAN-D2 \cite{Xian2019FVAEGAND2AF} & \cellcolor{red!10}76.1 & \cellcolor{red!10}\color{blue}57.1 & \cellcolor{red!10}65.2  & \cellcolor{green!10}75.6 & \cellcolor{green!10}63.2 & \cellcolor{green!10}68.9 & \cellcolor{blue!10}50.1 & \cellcolor{blue!10}37.8 & \cellcolor{blue!10}43.1 \\ 

SGMA \cite{SGMA}     & \cellcolor{red!10}87.1 & \cellcolor{red!10}37.6 & \cellcolor{red!10}52.5  & \cellcolor{green!10}71.3 & \cellcolor{green!10}36.7 & \cellcolor{green!10}48.5 & \cellcolor{blue!10}- & \cellcolor{blue!10}- & \cellcolor{blue!10}- \\


IIR \cite{IIR}     & \cellcolor{red!10}83.2 & \cellcolor{red!10}48.5 & \cellcolor{red!10}61.3  & \cellcolor{green!10}52.3 & \cellcolor{green!10}55.8 & \cellcolor{green!10}53.0 & \cellcolor{blue!10}30.4 & \cellcolor{blue!10}\color{blue}47.9 & \cellcolor{blue!10}36.8 \\

*E-PGN \cite{E-PGN}     & \cellcolor{red!10}83.5 & \cellcolor{red!10}52.6 & \cellcolor{red!10}64.6  & \cellcolor{green!10}61.1 & \cellcolor{green!10}52.0 & \cellcolor{green!10}56.2 & \cellcolor{blue!10}- & \cellcolor{blue!10}- & \cellcolor{blue!10}- \\


SELAR \cite{SELAR}       & \cellcolor{red!10}78.7 & \cellcolor{red!10}32.9 & \cellcolor{red!10}46.4  & \cellcolor{green!10}\color{red}76.3 & \cellcolor{green!10}43.0 & \cellcolor{green!10}55.0 & \cellcolor{blue!10}37.2 & \cellcolor{blue!10}23.8 & \cellcolor{blue!10}29.0 \\

\hline
\hline
ResNet-101 \cite{He2016DeepRL} & \cellcolor{red!10}66.7 & \cellcolor{red!10}40.1 & \cellcolor{red!10}50.1 & \cellcolor{green!10}59.5 & \cellcolor{green!10}52.3 & \cellcolor{green!10}55.7  & \cellcolor{blue!10}35.5 & \cellcolor{blue!10}28.8 & \cellcolor{blue!10}31.8 \\

\textbf{ResNet-101 with Implicit Attention} & \cellcolor{red!10}\textbf{74.1} & \cellcolor{red!10}\textbf{45.9} & \cellcolor{red!10}\textbf{56.8} & \cellcolor{green!10}\textbf{62.7} & \cellcolor{green!10}\textbf{54.5} & \cellcolor{green!10}\textbf{58.3} & \cellcolor{blue!10}\textbf{36.3} & \cellcolor{blue!10}\textbf{31.9} & \cellcolor{blue!10}\textbf{33.9} \\ 
\hline

\textbf{Our model (ViT)}
& \cellcolor{red!10}\textbf{{\color{blue}90.0}} & \cellcolor{red!10}\textbf{51.9} & \cellcolor{red!10}\textbf{{\color{blue}65.8}} & \cellcolor{green!10}\textbf{{\color{blue}75.2}} & \cellcolor{green!10}\textbf{{67.3}} & \cellcolor{green!10}\textbf{{\color{blue}71.0}} & \cellcolor{blue!10}\textbf{{\color{red}55.3}} & \cellcolor{blue!10}\textbf{{44.5}} & \cellcolor{blue!10}\textbf{{\color{blue}49.3}} \\

\textbf{Our model (ViT) with Implicit Attention}
& \cellcolor{red!10}\textbf{{89.9}} & \cellcolor{red!10}\textbf{53.7} & \cellcolor{red!10}\textbf{{\color{red}67.2}} & \cellcolor{green!10}\textbf{73.8} & \cellcolor{green!10}\textbf{{\color{red}68.6}} & \cellcolor{green!10}\textbf{{\color{red}71.1}} & \cellcolor{blue!10}\textbf{{\color{blue}54.7}} & \cellcolor{blue!10}\textbf{{\color{red}48.2}} & \cellcolor{blue!10}\textbf{{\color{red}51.3}} \\ 

\hline
\end{tabular} 

\begin{tablenotes}
\small
\item S, U, H denote Seen classes ($\mathcal{Y}^s$), Unseen classes ($\mathcal{Y}^u$), and the Harmonic mean, respectively. For each scenario, the best is in {\color{red}red} and the second-best is in {\color{blue}blue}. * indicates generative representation learning methods.
\end{tablenotes}
\end{table}


Table \ref{tab:Perfromance} illustrates a quantitative comparison between the state-of-the-art methods and the proposed method using two different backbones: (1) ResNet-101 \cite{He2016DeepRL} and (2) ViT \cite{ViT}. The baseline models performance without the employment of the SSL approach is also reported. The performance of each model is shown in \% in terms of Seen (S) and Unseen (U) classes and their harmonic mean (H). As reported, the classical ZSL models \cite{5206594,DeViSE,ConSE,ConSE,LATEM,ALE} show good performance in terms of seen classes. However, they perform poorly on unseen classes and encounter the bias issue, resulting in a very low harmonic mean. Among the classical approaches, \cite{ALE} performs the best on all the three datasets, as it overcomes the shortcomings of the previous models and considers the dependency between attributes. 
Among generative approaches, f-VAEGAN-D2 \cite{Xian2019FVAEGAND2AF} performs the best. Although f-CLSWGAN \cite{CLSWGAN} achieves the highest score on AWA2 unseen classes, it shows lower harmonic means on all the datasets than \cite{Xian2019FVAEGAND2AF}. As noticed, the first top scores for the AWA2 unseen classes accuracy are obtained by generative models \cite{CLSWGAN,Xian2019FVAEGAND2AF}, which we assume is because they include both seen and synthesised unseen features during the training phase. Moreover, attention-based models, such as \cite{SGMA,AREN} are the closest to our proposed model, perform better than the other models due to the inclusion of global and local representations. \cite{AREN} outperforms all reported models on the unseen classes of the CUB dataset, but still has low harmonic means on all the datasets. SGMA \cite{SGMA} performs poorly on both AWA2 and CUB, and it clearly suffers from the bias issue, where its performance on unseen classes is considered deficient compared to other models. Recent models such as SELAR \cite{SELAR} uses global maximum pooling as an aggregation method and achieves the best scores on CUB seen classes, but achieves low harmonic means. In addition, its performance is seen to be considerably impacted by the bias issue. 


\myparagraph{ResNet-101:} For a fair evaluation of the robustness and effectiveness of our proposed alternative attention-based approach, we consider the ResNet-101 \cite{He2016DeepRL} as one of our backbones, which is also used in prior related arts \cite{DeViSE,ALE,SAE,E-PGN,SELAR}. We have used the ResNet-101 backbone as a baseline model, where we only consider the global representation. Moreover, we also use this backbone with implicit attention, i.e. during training, we simultaneously impose a self-supervised image rotation angle prediction task for training the model. Note, for producing the results in Table \ref{tab:Perfromance}, we only use the images from the seen classes as set $\mathcal{A}$, which is used for rotation angle prediction task. As presented in Table \ref{tab:Perfromance}, our model with ResNet-101 backbone has performed inferiorly compared to our implicit and explicit variant, which will be discussed in the next paragraph. However, even with the ResNet-101 backbone, the contribution of our implicit attention mechanism should be noted, which provides a substantial boost to the model performance. For example, on AWA2, a considerable increment is observed on both seen and unseen classes, leading to a significant increase in the harmonic mean (i.e. $50.1\%$ to $56.8\%$). The performance of the majority of the related arts seems to suffer from bias towards the seen classes. We argue that our method tends to mitigate this issue on all the three datasets. Our method enables the model to learn the visual representations of unseen classes implicitly; hence, the performance is increased, and the bias issue is alleviated. Similarly, on the SUN dataset, although this dataset consists of $717$ classes, the proposed implicit attention mechanism illustrates the capability of providing ResNet-101 with an increase in the accuracy in terms of both seen and unseen classes, leading to an increase of $\approx 2$ points in the harmonic mean, i.e. from $31.8\%$ to $33.9\%$.

\myparagraph{Vision Transformer (ViT)}: We have used Vision Transformer (ViT) as another backbone to enable explicit attention in our model. Similar to the ResNet-101 backbone, we use the implicit attention mechanism with ViT backbone as well. During training, we simultaneously impose self-supervised image rotation angle prediction task for training the model. Here also we only use the images from the seen classes for image rotation angle task. As shown in Table \ref{tab:Perfromance}, consideration of explicit attention performs very well on all the three datasets and it outperforms all the previously reported results with a significant margin. Such results are expected due to the involvement of self-attention employed in ViT. It captures both the global and local features explicitly guided by the class attributes given during training. Furthermore, attention focuses to each element of the input patch embeddings after the image is split, which effectively weigh the relevance of different patches, resulting in more compact representations. Although explicit attention mechanism is seen to provide better visual understanding, the effectiveness of the implicit attention process in terms of recognising the image rotation angle is also quite important. It does not only improve the performance further but also reduces the bias issue considerably, which can be seen in the performance of the unseen classes. In addition, it allows the model via an implicit use of self-attention to encapsulate the visual features and regions that are semantically relevant to the class attributes. Our model achieves the highest harmonic mean among all the reported models on all the three datasets. In terms of AWA2, our approach scores the third highest accuracy on both seen and unseen classes, but the highest harmonic mean. Note that on AWA2 dataset, our model still suffers from bias towards seen classes. We speculate that is due to the lack of the co-occurrence of some vital and identical attributes between seen and unseen classes. For example, attributes \textit{nocturnal} in bat, \textit{longneck} in giraffe or \textit{flippers} in seal score the highest attributes in the class-attribute vectors, but rarely appear among other classes. However, on CUB dataset, this issue seems to be mitigated, as our model scores the highest harmonic mean (i.e. $H = 71.1\%$), where the performance on unseen classes is increased compared to our model with explicit attention. Finally, our model with implicit and explicit attention achieves the highest scores on classes on the SUN dataset, resulting in the best achieved harmonic mean. In summary, our proposed implicit and explicit attention mechanism proves to be very effective across all the three considered datasets. Explicit attention using the ViT backbone with multi-head self-attention is quite important for the good performance of the ZSL model. Implicit attention in terms of self-supervised pretext task is another important mechanism to look at, as it boosts the performance on the unseen classes and provides better generalisation.





\begin{figure}[!ht]
\begin{center}
\begin{tabular}{c c c c c}
\textbf{Original Images} & \multicolumn{2}{c}{\textbf{Attention Maps}} & \multicolumn{2}{c}{\textbf{Attention Fusions}} \\ 
& \multirow{2}{*}{Explicit Attention} & Implicit + & \multirow{2}{*}{Explicit Attention} & Implicit + \\
&  & Explicit Attention &  & Explicit Attention \\
\includegraphics[width=.19\textwidth, height=.15\textwidth]{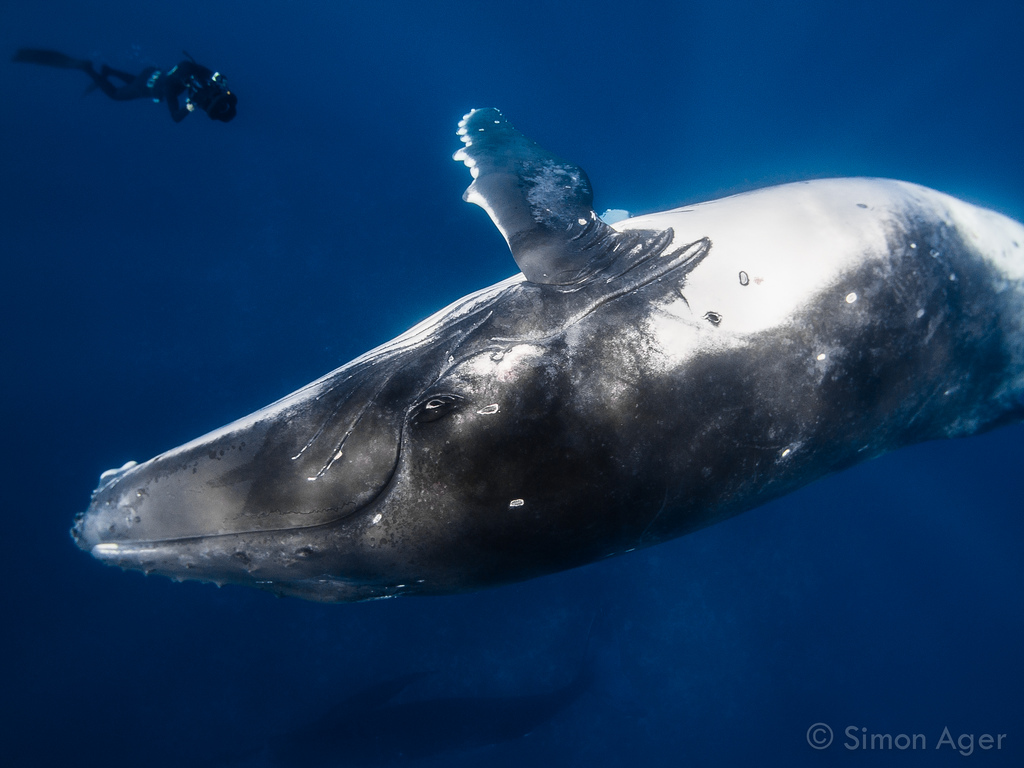}
& 
\includegraphics[width=.19\textwidth, height=.15\textwidth]{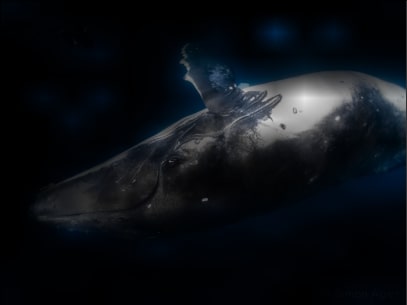}
& 
\includegraphics[width=.19\textwidth, height=.15\textwidth]{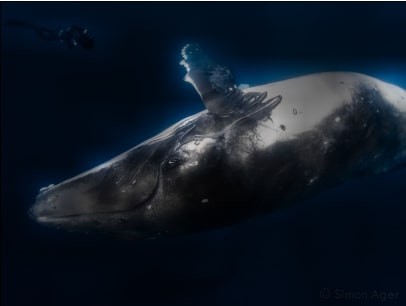}
& 
\includegraphics[width=.19\textwidth, height=.15\textwidth]{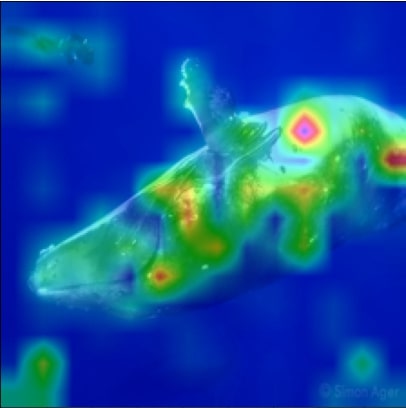}
& 
\includegraphics[width=.19\textwidth, height=.15\textwidth]{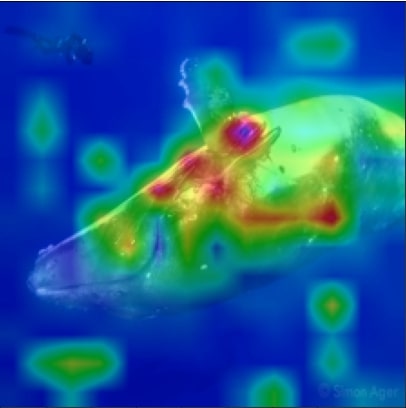} \\ \hline

    
\includegraphics[width=.19\textwidth, height=.15\textwidth]{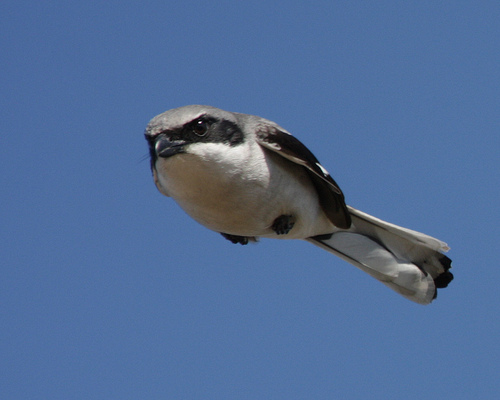}
& 
\includegraphics[width=.19\textwidth, height=.15\textwidth]{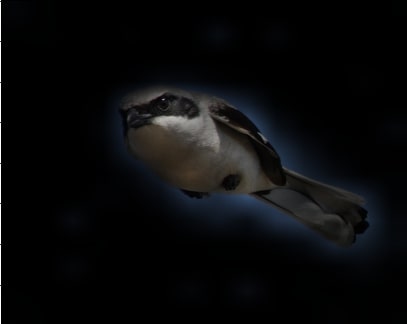}
& 
\includegraphics[width=.19\textwidth, height=.15\textwidth]{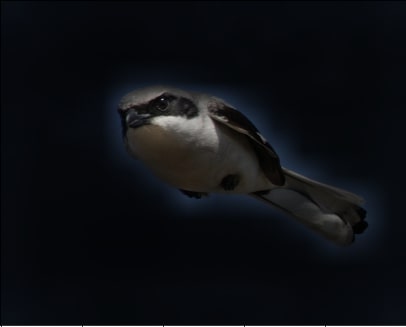}
& 
\includegraphics[width=.19\textwidth, height=.15\textwidth]{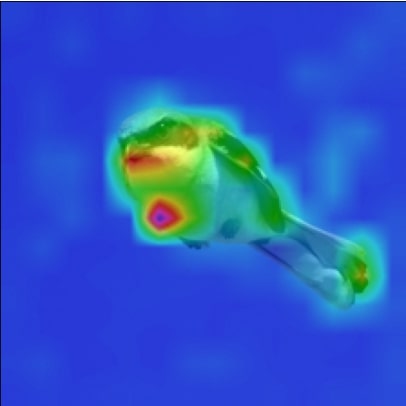}
& 
\includegraphics[width=.19\textwidth, height=.15\textwidth]{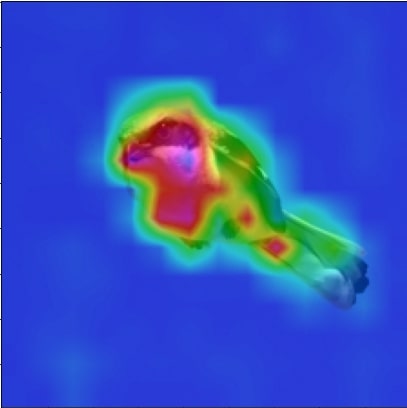} \\ 
\includegraphics[width=.19\textwidth, height=.15\textwidth]{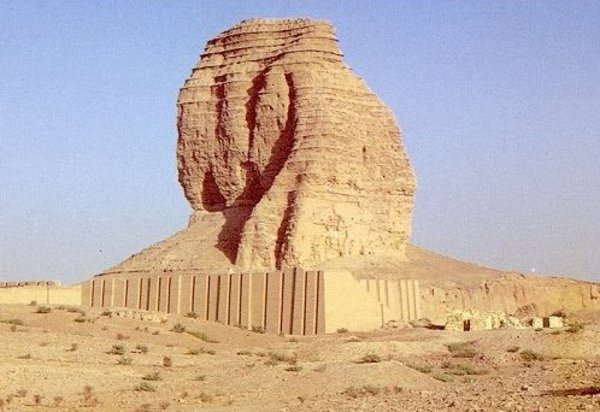}
& 
\includegraphics[width=.19\textwidth, height=.15\textwidth]{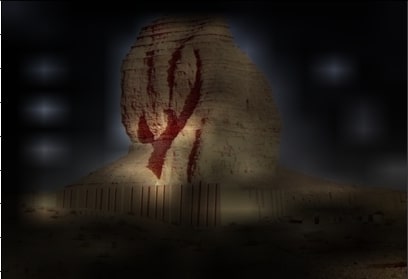}
& 
\includegraphics[width=.19\textwidth, height=.15\textwidth]{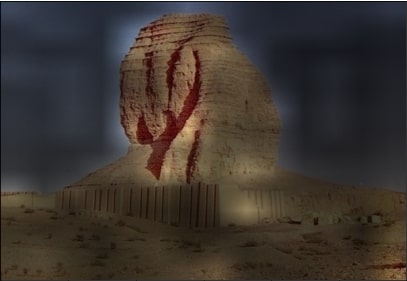}
& 
\includegraphics[width=.19\textwidth, height=.15\textwidth]{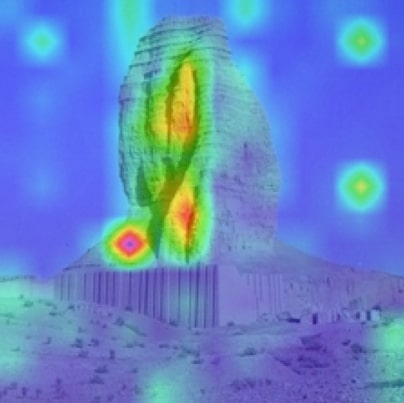}
& 
\includegraphics[width=.19\textwidth, height=.15\textwidth]{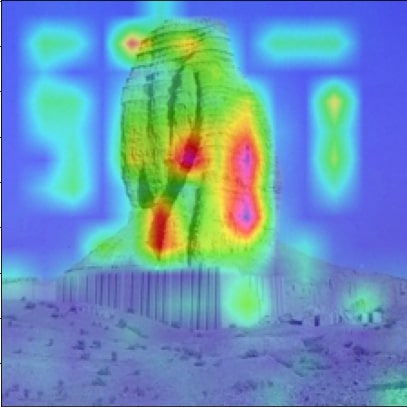} \\ 
\includegraphics[width=.19\textwidth, height=.15\textwidth]{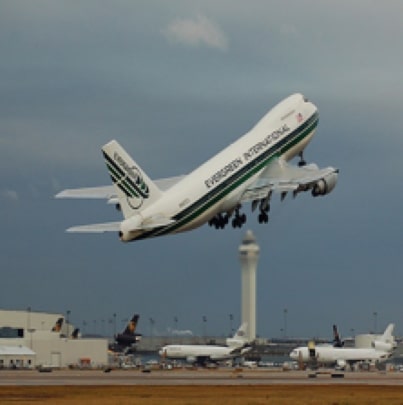}
& 
\includegraphics[width=.19\textwidth, height=.15\textwidth]{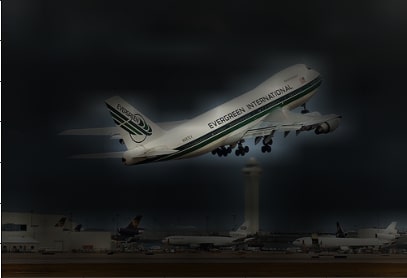}
& 
\includegraphics[width=.19\textwidth, height=.15\textwidth]{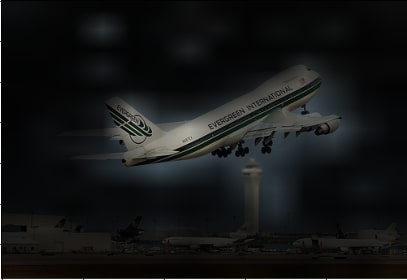}
& 
\includegraphics[width=.19\textwidth, height=.15\textwidth]{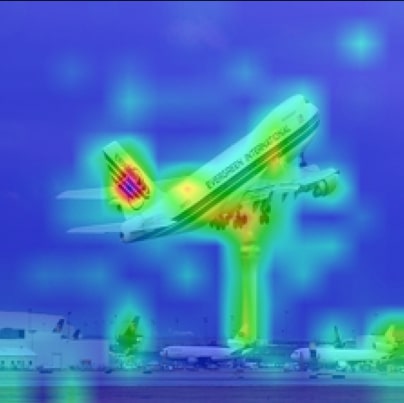}
& 
\includegraphics[width=.19\textwidth, height=.15\textwidth]{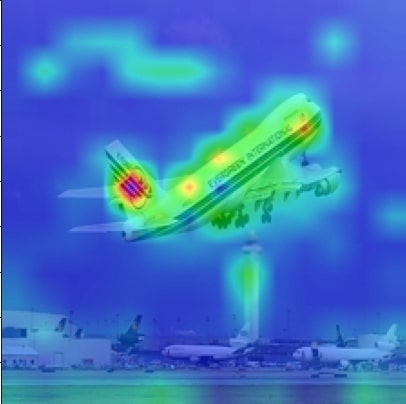}
\end{tabular}
\caption{Examples of implicit and explicit attention. First column: Original images, Second and Third: Attention maps without and with SSL, respectively, Four and Fifth: Attention Fusions without and with SSL, respectively. Our model benefits from using the attention mechanism and can implicitly learn object-level attributes and their discriminative features.}
\label{tbl:attentionMaps}
\end{center}
\end{figure}


\myparagraph{Attention Maps:} Fig. \ref{tbl:attentionMaps} presents some qualitative results, i.e. attention maps and fusions obtained by our proposed implicit and explicit attention-based model. For generating these qualitative results, we have used our model with explicit attention mechanism, i.e. we have used the ViT backbone. Attention maps and fusions are presented on four randomly chosen images from the considered datasets. Explicit attention with ViT backbone seems to be quite important for the ZSL tasks as it can perfectly focus on the object appearing in the image, which justifies the better performance obtained by our model with ViT backbone. Inclusion of implicit attention mechanism in terms of self-supervised rotated image angle prediction further enhances the attention maps and particularly focuses on specific image parts important for that object class. For example, as shown in the first row of Fig. \ref{tbl:attentionMaps}, our model with implicit and explicit attention mechanism focuses on both global and local features of the \textit{Whale} (i.e. water, big, swims, hairless, bulbous, flippers, etc.). Similarly, on the CUB dataset, the model pays attention to objects' global features, and more importantly, the discriminative local features (i.e. \textit{loggerhead shrike} has a white belly, breast and throat, and a black crown forehead and bill). For natural images taken from the SUN dataset, our model with implicit attention is seen to focus on the \textit{ziggurat} paying more attention to its global features. Furthermore, as in the \textit{airline} image illustrated in the last row, our model considers both global and discriminative features, leading to precise attention map that focuses accurately on the object.

\begin{table}[!ht]
\centering
\caption{Ablation performance of our model with ResNet-101 and ViT backbone on AWA2, CUB and SUN datasets. Here we use the training images from the seen classes as $\mathcal{S}$ and varies $\mathcal{A}$ as noted in the first column of the following table. S, U and PASCAL respectively denote the training images from the seen classes, test images from the unseen classes, and PASCAL VOC2012 training set images.}
\label{tab:ResNetPerfromance_Ablation}
\begin{tabular}{|c|c|c|c|c|c|c|c|c|c|c|c|} 
\hline
\multirow{2}{*}{\makecell{Source of Rotated\\Images ($\mathcal{A}$)}}  & \multirow{2}{*}{\makecell{Backbone\\(Implicit Attention)}} & \multicolumn{3}{c|}{\cellcolor{red!10}AWA2} & \multicolumn{3}{c|}{\cellcolor{green!10}CUB} & \multicolumn{3}{c|}{\cellcolor{blue!10}SUN} \\
\cline{3-10}
& & \cellcolor{red!10}S & \cellcolor{red!10}U & \cellcolor{red!10}H 
& \cellcolor{green!10}S & \cellcolor{green!10}U & \cellcolor{green!10}H 
& \cellcolor{blue!10}S & \cellcolor{blue!10}U & \cellcolor{blue!10}H \\
\hline
\multirow{2}{*}{S \& U} & ResNet-101
& \cellcolor{red!10}79.9 & \cellcolor{red!10}44.2 & \cellcolor{red!10}56.4 & \cellcolor{green!10}60.1 & \cellcolor{green!10}56.0 & \cellcolor{green!10}58.0 & \cellcolor{blue!10}35.0 & \cellcolor{blue!10}33.1 & \cellcolor{blue!10}33.7 \\
& ViT
& \cellcolor{red!10}87.3& \cellcolor{red!10}56.8 & \cellcolor{red!10}68.8 & \cellcolor{green!10}74.2 & \cellcolor{green!10} 68.9 & \cellcolor{green!10} 71.1& \cellcolor{blue!10}54.7 & \cellcolor{blue!10}50.0 & \cellcolor{blue!10}52.2 \\ \hline

\multirow{2}{*}{PASCAL} & ResNet-101
& \cellcolor{red!10}72.0 & \cellcolor{red!10}44.3 & \cellcolor{red!10}54.8 & \cellcolor{green!10}62.5 & \cellcolor{green!10}53.1 & \cellcolor{green!10}57.4 & \cellcolor{blue!10}35.6 & \cellcolor{blue!10}30.3 & \cellcolor{blue!10}33.1 \\
& ViT
& \cellcolor{red!10}88.1& \cellcolor{red!10}51.8 & \cellcolor{red!10}65.2 & \cellcolor{green!10}73.4 & \cellcolor{green!10} 68.0 & \cellcolor{green!10} 70.6& \cellcolor{blue!10}55.2 & \cellcolor{blue!10}46.3 & \cellcolor{blue!10}50.6 \\ \hline
\multirow{2}{*}{PASCAL \& U} & ResNet-101
& \cellcolor{red!10}75.1 & \cellcolor{red!10}46.5 & \cellcolor{red!10}57.4 & \cellcolor{green!10}62.9 & \cellcolor{green!10}54.4 & \cellcolor{green!10}58.4 & \cellcolor{blue!10}33.7 & \cellcolor{blue!10}32.7 & \cellcolor{blue!10}33.2 \\
&  ViT 
& \cellcolor{red!10}89.8& \cellcolor{red!10}53.2 & \cellcolor{red!10}66.8 & \cellcolor{green!10}73.02 & \cellcolor{green!10} 69.7 & \cellcolor{green!10} 71.3& \cellcolor{blue!10}53.9 & \cellcolor{blue!10}51.0 & \cellcolor{blue!10}52.4 \\ \hline

\multirow{2}{*}{PASCAL \& S} & ResNet-101
& \cellcolor{red!10}73.1 & \cellcolor{red!10}44.5 & \cellcolor{red!10}55.4 & \cellcolor{green!10}62.5 & \cellcolor{green!10}53.2 & \cellcolor{green!10}57.5 & \cellcolor{blue!10}36.6 & \cellcolor{blue!10}30.1 & \cellcolor{blue!10}33.1 \\ 
&  ViT
& \cellcolor{red!10}91.2& \cellcolor{red!10}51.6 & \cellcolor{red!10}65.9 & \cellcolor{green!10}73.7 & \cellcolor{green!10} 68.8 & \cellcolor{green!10} 71.1& \cellcolor{blue!10}54.19 & \cellcolor{blue!10}46.9 & \cellcolor{blue!10}50.9 \\
\hline
\end{tabular} 
\end{table}

\subsection{Ablation Study} Our ablation study evaluates the effectiveness of our proposed implicit and explicit attention-based model for the ZSL tasks. Here we mainly analyse the outcome of our proposed approach if we change the set $\mathcal{A}$ which we use for sampling images for self-supervised image angle prediction task during training. In Section \ref{sec:quant_results}, we have only used the seen images for this purpose; however, we have also noted important observation if we change the set $\mathcal{A}$. Note, here we can use any collection of images as $\mathcal{A}$, since it does not need any annotation regarding its semantic class, because in this case, the only supervision used is the class corresponds to image angle rotation which can be generated online during training. In Table \ref{tab:ResNetPerfromance_Ablation}, we present results on all three considered datasets with the above mentioned evaluation metric, where we only vary $\mathcal{A}$ as noted in the first column of Table \ref{tab:ResNetPerfromance_Ablation}. Note, in all these settings $\mathcal{S}$ remains fixed, and it is set to the set of images from the seen classes. In all the settings, we observe that explicit attention in terms of ViT backbone performs significantly better than the classical CNN backbone, such as ResNet-101. We also observe that the inclusion of unlabelled images from unseen classes (can be considered as transductive ZSL \cite{ALE}) significantly boosts the performance on all the datasets (see rows 1 and 3 in Table \ref{tab:ResNetPerfromance_Ablation}). Moreover, we also observe that including datasets that contain diverse images, such as PASCAL \cite{pascal} improve the performance on unseen classes and increase generalisation. 

\section{Conclusion}
This paper has proposed implicit and explicit attention mechanisms for solving the zero-shot learning task. For implicit attention, our proposed model has imposed self-supervised rotated image angle prediction task, and for the purpose of explicit attention, the model employs the multi-head self-attention mechanism via the Vision Transformer model to map visual features to the semantic space. We have considered three publicly available datasets: AWA2, CUB and SUN, to show the effectiveness of our proposed model. Throughout our extensive experiments, explicit attention via the multi-head self-attention mechanism of ViT is revealed to be very important for the ZSL task. Additionally, the implicit attention mechanism is also proved to be effective for learning image representation for zero-shot image recognition, as it boosts the performance on unseen classes and provides better generalisation. Our proposed model based on implicit and explicit attention mechanism has provided very encouraging results for the ZSL task and particularly has achieved state-of-the-art performance in terms of harmonic mean on all the three considered benchmarks, which shows the importance of attention-based models for ZSL task.


\section*{Acknowledgement}
This work was supported by the Defence Science and Technology Laboratory (Dstl) and the Alan Turing Institute (ATI). The TITAN Xp and TITAN V used for this research were donated by the NVIDIA Corporation.



%
\bibliographystyle{splncs04}
\bibliography{References}

\end{document}